\pdfoutput=1
\documentclass[11pt]{article}

\usepackage[]{emnlp2021}
\usepackage{array, makecell} 
\usepackage{times}
\usepackage{latexsym}
\usepackage{graphicx}
\usepackage{booktabs}

\usepackage[T1]{fontenc}
\usepackage[utf8]{inputenc}

\usepackage{microtype}

\usepackage{microtype}

\newcommand{\ignore}[1]{}

\title{A Bag of Tricks for Dialogue Summarization}

\newcommand*{\affaddr}[1]{#1} % No op here. Customize it for different styles.
\newcommand*{\affmark}[1][*]{{\small \textsuperscript{#1}}}
\newcommand*{\email}[1]{\texttt{#1}}

 \author{%
Muhammad Khalifa\affmark[2]\thanks{\hspace{0.1cm} Work done during an internship at Amazon.} , Miguel Ballesteros\affmark[1], 
Kathleen McKeown\affmark[1,3]\\
\affaddr{\affmark[1]Amazon AI, USA}\\
\affaddr{\affmark[2]Cairo University, Cairo, Egypt}\\
\affaddr{\affmark[3]Department of Computer Science, Columbia University, NY, USA}\\
\email{muhammad.e.khalifa@gmail.com} \\
\email{\{ballemig, mckeownk\}@amazon.com}

}

\date{}

\begin{document}
\maketitle
\begin{abstract}
Dialogue summarization comes with its own peculiar challenges as opposed to news or scientific articles summarization. In this work, we explore four different challenges of the task: handling and differentiating parts of the dialogue belonging to multiple speakers, negation understanding, reasoning about the situation, and informal language understanding. Using a pretrained sequence-to-sequence language model, we explore speaker name substitution, negation scope highlighting, multi-task learning with relevant tasks, and pretraining on in-domain data. Our experiments show that our proposed techniques indeed improve summarization performance, outperforming strong baselines.
\end{abstract}

\section{Introduction}
The nature of 
dialogue poses additional challenges to summarizers beyond what is required when processing structured, single-speaker documents \cite{zhu2006summarization}. 
%These challenges mainly arise from the fact that dialogues typically represent an interaction between multiple speakers, each having their own opinion and style. 
Given that dialogues typically represent an interaction between many speakers,
%As a result,
a summarizer model must keep track of the different lines of thoughts of individual speakers,
distinguish
salient from non-salient utterances, and finally produce a
%\kmcomment{I think it would be more striking if you said "produce a coherent monologue  from dialogue." or something along that line where you contrast that you are going from multi-speaker to single voice. However you want to say it. }
coherent, monologue summary of the dialogue. 
%In other words, having many speakers present in the dialogue, each expressing themselves, it becomes hard to decide which utterances from which speaker to take (or not take) into account while generating the final summary. Consequently, a single dialogue could have many possible summaries, each expressing a distinct point-of-view. 
%\muhcomment{Changed it with monologue-like}
%\kmcomment{I think you can make it just "monologue" and drop "like"}. 
%In addition,
%\kmcomment{something missing here. I added a phrase but check it is what you intended.}
%to learn
%a correct understanding of the scene, context is essential.
%both scene and context understanding are essential.

Dialogues usually include unfinished sentences where speakers were interrupted or repetitions, where a speaker expresses their thoughts more than once and possibly in different styles. Moreover, a single dialogue could touch on many topics without a clear boundary between the different topics. 
%In addition, dialogues can include co-reference to present or absent individuals. 
%Lastly, informal language can pose a challenge to methods trained on standard or formal text. 
All the aforementioned phenomena certainly add to the difficulty of the task \cite{zechner2000diasumm, zechner2002automatic,ChenY20}.

%Another challenge is represented by the \textit{subjectivity} of the task. With many speakers present in the dialogue, each expressing themselves, it becomes hard to decide which utterances from which speaker to take (or not take) into account while generating the final summary. Consequently, a single dialogue could have many possible summaries, each expressing a different point-of-view. 

Our work focuses on SAMSum \cite{gilwa2019}, which is a dialogue summarization dataset comprised of \textasciitilde 16K everyday dialogues with their human-written summaries. As our backbone model, we use BART \cite{bartlewis2020}, a state-of-the-art pretrained encoder-decoder language model that is suitable for sequence-to-sequence tasks. Table ~\ref{tab:example} shows an example of 
a summary generated using BART \cite{bartlewis2020}, fine-tuned on SAMSum. Clearly, a level of reasoning is required to make sense of the conversation, which BART fails to do and therefore produces an incorrect summary.

We propose a combination of techniques to tackle a set of dialogue summarization challenges. The first challenge is having \textbf{\textit{multiple speakers}} (generally, more than 2), where it becomes harder for the model to keep track of different utterances and determine their saliency. The second challenge is \textbf{\textit{multiple negations}}, which is thought by \newcite{ChenY20} to pose some difficulty to dialogue understanding.  The third of these challenges is \textit{\textbf{reasoning}}, where the model is required to reason about the dialogue context, and infer information that is not explicitly expressed. The last challenge is \textbf{\textit{informal language}}. Since we focus on random, everyday conversations, these are usually filled with non-standard language (abbreviations, social media terms, etc.).
%Figure ~\ref{fig:challenges} shows the different challenges associated with the task along with the four techniques we propose to handle them. 

%\kmcomment{Wording could be better here. I'm proposing some changes}
The contributions in this work are:
\begin{itemize}
     \item We propose a set of 
     novel techniques to address four dialogue summarization challenges: multiple speakers, negation, reasoning and informal language. Our techniques include 
   %KMFinal - I belive "names" should be singular. It's used as a mass noun here.  
 %    names 
     name substitution, negation scope highlighting, multi-task learning with relevant tasks, and pretraining on in-domain corpora.
     
     \item We show impressive improvements on the summarization performance using three of these, outperforming very strong baselines.
 \end{itemize} 
 
\section{Related Work}
Early work on dialogue summarization focused more on extractive than abstractive techniques for summarization of meetings
\cite{murray2005extractive,riedhammer2008keyphrase} or random conversations \cite{murray2007term}. In the context of meeting summarization, \newcite{shang-etal-2018-unsupervised} proposed an unsupervised graph-based sentence compression approach for meeting summarization on the AMI \cite{mccowan2005ami} and ICSI \cite{janin2003icsi} benchmarks. 
\newcite{goo2018abstractive} leveraged hidden representations from a dialogue act classifier through a gated attention mechanism to guide the summary decoder. 
 
More recently, \newcite{gilwa2019} proposed SAMSum, a benchmark for abstractive everyday dialogue summarization.
\newcite{zhao2020improving} modeled dialogues  using a graph structure of words and utterances and summaries are generated using a graph-to-sequence architecture. \newcite{ChenY20} proposed a multi-view summarization model, where views can include topic or stage. They also pointed out to seven different challenges to dialogue summarization and analysed the effect each challenge can have on summarization performance using examples from SAMSum.
%\kmcomment{Need a list of contributions. I am also wondering if this paper is going to address each of the challenges you present. If you introduce them, it should. }
%\muhcomment{Not sure if each technique is a contribution? Or should just say proposed four techniques}
%\kmcomment{I think it's OK to put them in one as you have done. }

\begin{table}[]
\footnotesize
    \centering
    \begin{tabular}{l}
    \toprule
    \textbf{Dialogue:} \\
    \hline 
    Orion: I miss him :( \\ 
    Cordelia: Need i remind you that he cheated \\ on you? You deserve alot better than that \\
    Orion: ...what? oh, right noo - im talking  about \\ my rat  ... he died \\ 
    %Cordelia: SMITHERS IS DEAD???!!!!!\\
    ... \\
    
    \textbf{Vanilla BART Output:} \\
    \hline
    Orion's rat died. He cheated on her. \\
    %\textbf{Our Model Output:} \\
    %\hline 
    % Orion's rat died and he misses him.
    \textbf{MTL BART output:} \\
    \hline
    Orion's rat died and he misses him. \\
    \textbf{Reference:} \\
    \hline
    Orion is grieving after the death of her rat.\\
    \bottomrule
    \end{tabular}
    \caption{Example from SAMSum \cite{gilwa2019} of a dialogue and its generated summaries using two BART models: vanilla and multi-tasked. The summary generated by the vanilla model indicates that the rat is the cheater, pointing to a lack of commonsense reasoning on the model side. The output of our multi-tasked model (section ~\ref{sec:reasoning}) clearly shows better understanding of the dialogue. }
    \label{tab:example}
\end{table}

%\section{Related Work}
%\input{related}

% \begin{table*}[h!]
%     \centering
%     \begin{tabular}{l|l}
%     \hline
%     \textbf{Challenge}    & \textbf{Technique}  \\
%     \hline
%     Reasoning   & MTL on relevant tasks. \\ 
%     \hline
%     Informal Language  & pretraining on in-domain data. \\ 
%     %Style Transfer 
%     \hline
%     Negation    & Negation scope highlighting \\ 
%     \hline
%     Co-reference + Multiple Participants    & Speaker named replacement \\ 

%     \hline

%     \end{tabular}
%     \caption{Challenges and Proposed Techniques}
%     \label{tab:challenges}
% \end{table*}

% \begin{figure*}[h!]
%     \centering
%     \includegraphics[width=11cm]{dialogue-challenges.pdf}
%     \caption{Challenges and proposed techniques to handle each challenge.}
%     \label{fig:challenges}
% \end{figure*}

%\section{Related Work}
\section{Challenges}
We now present our four techniques for dialogue summarization: name substitution (section~\ref{sec:mult-speakers}), negation 
%KMFinal - "scopes should be singular
%scopes 
scope highlighting (section \ref{sec:negation}), multi-task learning on common sense tasks (section \ref{sec:reasoning}), and pretraining on an in-domain dialogue corpus (section \ref{sec:inf-language}).

%We hypothesize that BART has seen little dialogue data during pretraining, which makes it relatively less adaptable to the unstructured nature of dialogue input. Therefore, in order to improve dialogue summarization performance, we aim to adapt part dialogues through few techniques:

% \begin{itemize}
%     \item Continue pretraining BART on a dialogue-related corpus.
%     \item Continue pretraining BART but add dialogue-specific pretraining objectives.
    
% \end{itemize}

\subsection{Experimental Setup}
For all our experiments, we use BART large architecture \cite{bartlewis2020}.\footnote{For fine-tuning, we use ADAM optimizer with a learning rate of $0.00002$ and label smoothing with $\alpha=0.1$.} All our experiments are run using fairseq \cite{fairseq2019}.

\subsection{Baselines}
We compare our techniques to two summarization baselines:

\begin{itemize}
    \item \textbf{Vanilla BART}: Fine-tuning the original BART large checkpoint model on SAMSum.
    \item \textbf{Multi-view Seq2Seq} \cite{ChenY20} : This is based on BART, as well, but during the summarization, the model considers multiple views, each of which defines a certain structure for the dialogue. We compare to their best model which combines topic and stage views.
\end{itemize}

%\miguelcomment{you need to add a section with baseline and explanation of baseline results. then explain that we observed and others have observed many rooms for improvement and certain challenges to improve, then...}
%\muhcomment{I think this better be done at the related work section}

%\miguelcomment{I would add a section with "Challenges", each challenge will be a Subsection and then the technique to solve it as part of it with results}

\subsection{Multiple Speakers}
\label{sec:mult-speakers}
%\kmcomment{This is a nice description of what you do. If you could use some of this wording for the contributions that would be good. }
We hypothesize that uncommon (less frequent in the original pretraining data) or new names could be an issue to a pretrained model, especially if such names were seen very few times, or not at all, during pretraining. Such issues could specifically show up in multi-participant conversations, and could introduce co-reference issues when generating the summary. As a simple technique to alleviate this, we preprocess SAMSum by replacing speaker names with more common, frequent names, ones that the model is more likely to have seen during pretraining. Since we are dealing with English dialogue summarization, we use a list\footnote{https://www.ssa.gov/oact/babynames/decades/century.html} of common English names and replace each speaker name with a randomly sampled same-gender name from this list. Since the name list is divided by gender (male or female), we use \texttt{gender guesser}\footnote{https://github.com/lead-ratings/gender-guesser} to replace with a same-gender name. To avoid modifying the ground truth summaries and to ensure a fair comparison with other models, the original name is replaced back into the generated summary before evaluation.

\begin{figure*}
    \centering
    \includegraphics[width=\linewidth, height=3.4cm]{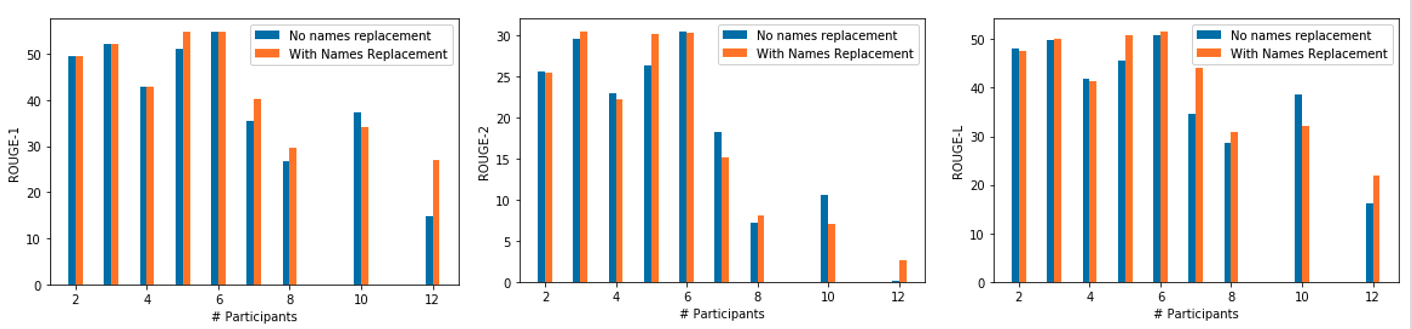}
    \caption{ROUGE values against the number of participants per dialogue on the development set of SAMSum. Performance boost is more clear in dialogues with more participants }
    \label{fig:mask-names}
\end{figure*}

Table~\ref{tab:mask-names} compares the performance of this technique to fine-tuning BART on the original SAMSum data. We observe ROUGE improvements on both validation and test sets of SAMSum. In addition, we perform an analysis of the performance with respect to the number of participants per dialogue. Figure~\ref{fig:mask-names} plots the summarization performance against the number of speakers. We can see that conversation with more participants (7, 8, 12) exhibit higher ROUGE boost than conversations with fewer speakers (2, 3, 4). In other words, we observe that the more participants in the summary, the more effect this technique has. Notably, the average number of speakers per dialogue in SAMSum is only \textasciitilde 2.4. 
and we expect name substitution to work even better with datasets that have many more speakers per dialogue.

\begin{table*}[]
\footnotesize
\centering
\begin{tabular}{|c|ccc|ccc|}
\hline
\textbf{Data} &  \multicolumn{3}{c|}{\textbf{Val}} & \multicolumn{3}{c|}{\textbf{Test}} \\
 \hline
 & \textbf{R-1}     & \textbf{R-2}     & \textbf{R-L}    & \textbf{R-1}     & \textbf{R-2}     & \textbf{R-L}     \\
 \hline
 
SAMSum &  49.22 &	26.47 &	47.80 &	48.65 &	25.20 &	47.08 \\
 SAMSum + name substitution & \textbf{49.98} &	\textbf{26.50} &	\textbf{48.48} & \textbf{49.09} & \textbf{25.91} & \textbf{47.87} \\ 
 \hline

\end{tabular}
\caption{ROUGE-1, ROUGE-2, and ROUGE-L on SAMSum with and without names substitution. Results are shown on the validation and test splits from \cite{gilwa2019}. %\miguelcomment{explain how test and validation numbers were obtained}
}
\label{tab:mask-names}
\end{table*}

\subsection{Negation Understanding}
\label{sec:negation}
\newcite{ChenY20} argue that negations represent a challenge for dialogues. We experiment with marking negation scopes in the input dialogues before feeding them to BART. To do that, we fine-tune a RoBERTa base model on the CD-SCO dataset
%\footnote{https://www.clips.uantwerpen.be/sem2012-st-neg/\#Datasets} 
from SEM Shared Task 2012 for negation scope prediction \cite{morante2012sem}. Then, we mark negation scope using two  designated special tokens to mark the start and the end of the negation scope. For example, the sentence \textit{``I don't know what to do''} becomes \textit{``I don't \texttt{\textless   NEG\textgreater}  know what to do \texttt{\textless \textbackslash NEG\textgreater}''} after negation scope highlighting. We initialize the embeddings of the special tokens \texttt{\textless   NEG\textgreater} and \texttt{\textless \textbackslash NEG\textgreater} randomly.

Results are shown in Table~\ref{tab:nagation-results}. While we expected to see a performance boost due to negation scope highlighting, we actually saw a performance drop except on ROUGE-L on the test set. To understand why, we investigate the negation challenge dialogues put together in \cite{ChenY20}. We found that in all examples, negation did not seem to be a problem, and that BART was able to handle multiple negations very well. Therefore marking negation scopes could have introduced unneeded noise into the model, causing the observed performance drop.

\begin{table*}[]
\footnotesize
\centering
\begin{tabular}{|c|ccc|ccc|}
 \hline

\textbf{Data} &  \multicolumn{3}{c|}{\textbf{Val}} & \multicolumn{3}{c|}{\textbf{Test}} \\
 %\hline
 \cline{2-7}
 & \textbf{R-1}     & \textbf{R-2}     & \textbf{R-L}    & \textbf{R-1}     & \textbf{R-2}     & \textbf{R-L}     \\
 \hline
 
 Original SAMSum &  \textbf{49.22} &	\textbf{26.47} &	\textbf{47.80} &	\textbf{48.65} &	\textbf{25.20} & 47.08 \\
 SAMSum + negation scope marked & 48.61&	25.45&	47.82&	48.59&	24.96&\textbf{	47.32} \\ 
 \hline

\end{tabular}
\caption{Summarization performance on SAMSum when highlighting negation scope.}
\label{tab:nagation-results}
\end{table*}

\subsection{Reasoning}
%\kmcomment{I think it would be very helpful if you could refer to an example of this. Is there one in the examples you have included? }
%Correct 
Reasoning is often necessary for dialogue summarization \cite{ChenY20}, especially in cases where there is missing information or implicit assumptions regarding the situation. Unfortunately, it is difficult for the model to learn to conduct such reasoning by relying only on the reference summaries (this difficulty is 
exacerbated by the fact that SAMSum is of a relatively small size). Multi-task learning (MTL) enables knowledge transfer across relevant tasks. For instance \newcite{li2019keep} improved their summarization performance by jointly learning summarization and topic segmentation. Also, \newcite{konar2020ana} improved commonsense reasoning through multi-task learning on relevant datasets. Similarly, we propose to simultaneously learn summarization and other reasoning-based tasks.

\label{sec:reasoning}
%\kmcomment{I think this should start with more of an explanation of why we tried this. An example of where BART misses the point, which is the end of the story, would help. }
More specifically, we jointly fine-tune BART on the following tasks :
%\kmcomment{If you had an example that showed that the point of the story is hwat you want to include in the summary then I think it would be more clear why short story ending prediction could be helpful. }
\begin{itemize}
    \item \textbf{Short Story Ending Prediction}: this task could be helpful as predicting story ending requires intuitive understanding of the events. Also, conversation endings could be essential to understand the point of the dialogue (See examples 1 and 2 in Table \ref{App:tab-reasoning-examples} in the Appendix~\ref{App:reasoning}). %We hypothesize that such understanding could help with dialogue understanding and summarization. 
    We use the ROC stories dataset \cite{mostafazadeh-etal-2016-corpus}.

\item \textbf{Commonsense Generation}: Generative commonsense reasoning \cite{LinSZZBCR20} is a task involving generating an everyday scenario description given basic concepts. We assume such task could help the model reason more about conversations, which is certainly needed in many dialogues (see example 3 in Table \ref{App:tab-reasoning-examples} in Appendix~\ref{App:reasoning}).

\item \textbf{Commonsense Knowledge Base Construction}: The task here is to generate relation triplets similar to \cite{comet19}. More specifically, we train our model to predict relation objects given both relation and subject. We use ConceptNet \cite{liu2004conceptnet}.
\end{itemize}
Table~\ref{tab:mtl-results} shows the summarization performance after multi-task fine-tuning of BART. We also show the results of combining ROC and CommonGen with SAMSum. It is clear that MTL gives a performance boost in almost all cases, outperforming the vanilla BART and the Multi-view SS baseline on both the development and test sets. It is worth noting that due to the small size of both validation and test splits (\textasciitilde 800 dialogues), it is difficult to test the  statistical significance of these results.

%\kmcomment{Since, as I remember, you can't show significance, perhaps a sentence that the dataset is small and thus, difficult to show significance? }

\begin{table*}[h!]
\footnotesize
\centering
\begin{tabular}{|l|lll|lll|}
 \hline

\textbf{Tasks} &   \multicolumn{3}{c|}{\textbf{Val}} & \multicolumn{3}{c|}{\textbf{Test}} \\
 \hline
 & \textbf{R-1}     & \textbf{R-2}     & \textbf{R-L}    & \textbf{R-1}     & \textbf{R-2}     & \textbf{R-L}     \\
  \cline{2-7}
 
 SAMSum &   49.22 &	26.47 &	47.80 &	48.65 &	25.20 & 47.08 \\
 
 Multi-view SS \cite{ChenY20} & - & - & - & 49.30 & 25.60 & 47.70 \\ 

\hline
 SAMSum + ROC & \textbf{50.44} &	26.63 &	\textbf{48.78} & 49.31 & 26.18 & \textbf{48.18} \\

 SAMSum + CommonGen &   50.09 &	\textbf{26.86} &	48.73 & 49.12 &	25.76 &	47.71 \\
 SAMSum + ConceptNet &   49.70 &	26.65 &	48.26 & 49.03 & 25.71 &	47.92 \\

 SAMSum + ROC + CommonGen &  49.22	& 26.47 &	47.80 & 	\textbf{49.45} &	\textbf{26.20} &	47.93 \\
\hline
\end{tabular}
\caption{Summarization performance on SAMSum when fine-tuning BART with multi-task learning of Commonsense generation (CommonGen), Knowledge Base Construction (ConceptNet), and Story Ending completion (ROC).}
\label{tab:mtl-results}
\end{table*}

\begin{table*}[h!]
\footnotesize
\centering
\begin{tabular}{|l|lll|lll|}
\hline
 \textbf{Pretraining Corpus} &  \multicolumn{3}{c|}{\textbf{Val}} & \multicolumn{3}{c|}{\textbf{Test}} \\
 \hline
 & \textbf{R-1}     & \textbf{R-2}     & \textbf{R-L}    & \textbf{R-1}     & \textbf{R-2}     & \textbf{R-L}     \\
 \cline{2-7}
 
 Original BART &   49.22 &	26.47 &	47.80 &	48.65 &	25.20 & 47.08 \\
  Multi-view SS \cite{ChenY20} & - & - & - & \textbf{49.30} & 25.60 & 47.70 \\ 
\hline 
PersonaChat (entities, pronouns, tfidf) & 50.07 &	26.81 &	48.68 & 48.66 &	25.26 &	47.39 \\ PersonaChat (span masking) & 49.59 &26.11 & 47.97 &		48.88 &	25.52 &	47.63 \\
PersonaChat (word masking) & \textbf{50.17} & \textbf{26.99} &	\textbf{48.95} &		49.22 &	25.64 &	\textbf{47.90} \\
\hline
PersonaChat + Reddit (entities, pronouns, tfidf)& 49.64 &	26.31 &	48.38 & 48.43 &	25.09 &	47.23 \\
PersonaChat + Reddit (span masking) & 49.43 &	25.92 &	48.00 & 49.20 & \textbf{25.87} & 47.74 \\
PersonaChat + Reddit (word masking) & 49.12 &	26.03 &	47.84 & 48.99 & 25.52 & 47.63 \\

 \hline
\end{tabular}
\caption{Summarization performance on SAMSum when BART is pretrained on an in-domain corpus. We also include results when using additional dialogue-specific pretraining objectives (See Appendix ~\ref{Appendix:Pretraining-Obj}).}
\label{tab:pretrain-result}
\end{table*}

\begin{table*}[h!]
\footnotesize
\centering
\begin{tabular}{|l|lll|lll|}
 \hline
\textbf{Tasks} & \multicolumn{3}{c|}{\textbf{Val}} & \multicolumn{3}{c|}{\textbf{Test}} \\
 \hline
 & \textbf{R-1}     & \textbf{R-2}     & \textbf{R-L}    & \textbf{R-1}     & \textbf{R-2}     & \textbf{R-L}     \\
 \cline{2-7}
  Original BART &   49.22 &	26.47 &	47.80 &	48.65 &	25.20 & 47.08 \\
  Multi-view SS \cite{ChenY20} & - & - & - & 49.30 & 25.60 & 47.70 \\ 
%\hline 
%PersonaChat (word masking) & 50.17 & 26.99 &	48.95 &		49.22 &	25.64 &	47.90 \\

 \hline
 SAMSum + ROC & 50.44 &	26.63 &	48.78 & 49.31 & \textbf{26.18} & \textbf{48.18 }\\
 \hline
Pretraining + MTL(SAMSum, ROC) &  \textbf{50.48 }&	\textbf{27.25} &	\textbf{48.90} &	\textbf{49.34} &	25.54 &	47.88
 \\

Pretraining + MTL(SAMSum, ROC, CommonGen) &   50.29 &	27.21 &	49.05 &	\textbf{49.34} &	25.81 &	47.85 \\
 \hline
 
Pretraining + MTL(SAMSum, ROC) + name substitution & 49.97 &26.94 & 48.88 & 48.87 & 25.70 & 47.72 \\

\hline
 
\end{tabular}
\caption{Summarization performance on SAMSum when BART is pretrained on an in-domain corpus and then fine-tuned in a multi-task fashion.}
\label{tab:pretrain-mtl}
\end{table*}

\subsection{Informal Language}
\label{sec:inf-language}
We hypothesize that pretrained language models, BART in our case, are not well-adapted to the dialogue domain. Therefore, we 
%KM since you actually do it, you don't need "propose"
%propose 
adapt BART to dialogue inputs by further pretraining of BART on a dialogue corpus and with dialogue-specific objectives. \footnote{Our proposed dialogue-specific pretraining objectives are explained in Appendix~\ref{Appendix:Pretraining-Obj}.}

\subsubsection{Pretraining Corpora}
We consider the following 2 corpora for further pretraining of BART: PersonaChat (140K utterances) \cite{DBLP:conf/acl/KielaWZDUS18}, and a collection of 12M Reddit comments. We experiment with both whole word masking and span masking (masking random contiguous tokens). Our experimental setup is described in the Appendix in section~\ref{App:pretrain-settings}. 
%Ubuntu Corpus \cite{DBLP:conf/sigdial/LowePSP15},

% \begin{table}[]
%     \centering
%     \begin{tabular}{|c|c|c|}
%      \hline
%       \textbf{Dataset}  & \textbf{\# Dialogues} & \textbf{\# Utterances} \\
%         \hline
%         PersonaChat & 8.9K & 140K \\
%         Reddit & - & 12M \\
%         %Ubuntu & 930K & 7M  \\

%     \hline
%     \end{tabular}
%     \caption{Caption}
%     \label{tab:my_label}
% \end{table}

Table~\ref{tab:pretrain-result} shows the results of fine-tuning BART pretrained on dialogue corpora.\footnote{We experimented with pretraining only on Reddit, but found it to perform worse.} The best model (PersonaChat, word masking) outperforms the vanilla BART on all metrics and the Multiview SS baseline on test set ROUGE-2 and ROUGE-L. We can see that in general, BART pretrained on PersonaChat is better than pretraining on both PersonaChat and Reddit, which is surprising since more pretraining data usually means better performance. This could be explained by the dissimilarity between Reddit comments and the dialogues in SAMSum. We can also see that whole word masking performs slightly better than span masking. Based on these results, it is obvious that further pretraining on in-domain corpora can be helpful when dealing with inputs of special nature such as dialogues. 

Also, we can see that pretraining using dialogue-specific objectives is performing well (on either PersonaChat only or with Reddit), and even outperforming random span masking on the validation set. This certainly shows that task-specific pretraining could be beneficial. 

At last, we combine pretraining with MTL by fine-tuning a pretrained model in a multi-task learning fashion. Table~\ref{tab:pretrain-mtl} compares this to separate pretraining and MTL. We can see that pretraining on PersonaChat and fine-tuning on both SAMSum and ROC gives the best performance over the validation set, outperforming all other settings. 
On the test set, it is performing very well but slightly outperformed by multi-tasking with ROC in both ROUGE-2 and ROUGE-L. Lastly, we combine named substitution with the best model here and the results are also shown in Table~\ref{tab:pretrain-mtl}. We observe that name substitution does not give a performance boost when used in combination with pretraining and MTL.

\section{Conclusion}
%\kmcomment{One natural question is why you don't try all methods in combination}
%\muhcomment{We already combine MTL and pretraining, I will try to run another exp with name substitution as well}
%\kmcomment{We should think about how to do a better conclusion. For example, might want to highlight that things that seemed like problems were not actually problems. Or somehow highlight that negative results lead us to conclude that summarization on this dataset is a hard task and say why.}
In this paper, we explored different techniques to improve dialogue summarization performance by addressing different challenges to the task individually. The proposed techniques included name substitution, negation scope highlighting, multi-task learning with relevant tasks, and pretraining on in-domain corpora. On one hand, our experiments on three challenges showed the effectiveness of our proposed techniques which outperformed strong baselines on the task. On the other hand, our proposed technique to handle multiple negations performed poorly and by analyzing the outputs on negation-intensive dialogues, we found that multiple negations do not represent a challenge for dialogue summarization systems.

\section{Ethics Discussion}
\label{ethics}
We refer to Section \ref{sec:mult-speakers}, where we explain how we aid the model with name substitution using more common names (common here means more frequent in the pre/training data, and not by any preconception to us or any other entity). 
%This replacement is mapped back in the summarized output by the name that appears in the dialogue as it is just a way to test whether the model would do better with names that it has seen more often. 
As explained above, we are using a list of the most common names in American English, which is divided in feminine and masculine names. We therefore use \texttt{gender guesser} to ensure that the pronouns in the dialogue co-refer correctly with the replaced names. It is however worth mentioning that even if the character in the dialogue is non binary and/or the pronouns used in the dialogue are \textit{they/them}, our approach would work given that the replaced name would still co-refer with those pronouns and the name that is being replaced. We however hope to work in the future with datasets and list of names that contain non-binary gender.

\bibliographystyle{acl_natbib}
\bibliography{ref}

\appendix
\clearpage

%\section{Name Substitution}

% \begin{figure*}[t!]
%     \centering
%     \includegraphics[width=\linewidth, height=3.5cm]{names-replacement-validation}
%     \caption{ROUGE values against the number of participants per dialogue on the development set of SAMSum. Performance boost is more clear in dialogues with more participants }
%     \label{fig:mask-names}
% \end{figure*}

\section{Reasoning}
\label{App:reasoning} 
Here we show examples from SAMSum validation set where both story end understanding and reasoning about the situation are essential for correct summarization. Rows (1) and (2) in Table~\ref{App:tab-reasoning-examples} are examples of a dialogue where the main point of the conversation is only known in the last utterance. Consequently, we hypothesize that learning to predict story endings could teach the model to focus more on dialogue endings. 

Row (3) is an example of a situation that requires high-level commonsense reasoning. Given the information in the dialogue, it is \textit{very difficult} for the model to infer that the conversation is about a marriage proposal. Through our error analysis, we find that incorrect or incomplete reasoning is a major source of error in summarization. For example the output of vanilla BART on this dialogue is: "Colin congratulates Patrick on his girlfriend", which shows that the model clearly misses the point. Our best MTL model, on the other hand, produces "Patrick is over the moon because she said yes.", which is certainly better than vanilla BART.
\begin{table*}[thb]
\footnotesize
\centering\renewcommand\cellalign{lc}
\setcellgapes{3pt}\makegapedcells
\begin{tabular}{|l|c|l|} \hline
\textbf{\#} & \textbf{Dialogue} & \textbf{Reference} \\ \hline

\makecell{1} & 
\makecell{
\textbf{Keith:} Meg, pls buy some milk and cereals, \\
    I see now we've run out of them \\
    \textbf{Megan:} hm, sure, I can do that \\
    \textbf{Megan:} but did you check in the drawer next to \\
    the fridge? \\
    \textbf{Keith:} nope, let me have a look \\
    \textbf{Keith:} ok, false alarm, we have cereal and milk :D} 
    & 
    \makecell{Megan needn't buy milk and cereals. \\ They're in the drawer next to the fridge.}
    \\
    \hline 
    \makecell{2} & 
    \makecell{
    \textbf{Taylor:} I have a question!!\\
    \textbf{Isabel:} Yes? \\
    \textbf{Taylor:} Why haven’t you introduced me even once\\
    your bf to me? \\
    \textbf{Taylor:} All of my friends’ daughters bring their bfs\\
    and introduced them. \\
    \textbf{Taylor:} You know I’m such a cool mum. \\
    I won’t make him stressful. \\
    \textbf{Taylor:} Just bring him. \\
    \textbf{Isabel:} Because mum...I haven’t had any! \\
    }
    & 
    \makecell{
    Taylor wants to meet Isabel's \\boyfriend but she has never had any.
    }
    \\ \hline
    
    \makecell{3} 
    &
    \makecell{\textbf{Colin:} DUUDE, congrats! \\ \textbf{Patrick:} Thanks! \\
    \textbf{Patrick:} She said yes, I'm over the moon! \\ 
    \textbf{Colin:} Lucky guy \\}
    & 
    \makecell{
    Patrick's girlfriend accepted his proposal.
    } \\
    \hline 
    \end{tabular}
 \caption{Sample dialogues from SAMSum \cite{gilwa2019} that require reasoning for correct understanding/summarization.}
    \label{App:tab-reasoning-examples}
\end{table*}

\section{In-domain pretraining}

\subsection{Experimental Settings}
\label{App:pretrain-settings}
We continued pretraining BART for 50K gradient update steps with batch size of 1024 tokens and a learning rate of $0.00001$. We use $p_{mask}=0.3$ and for span masking, we sample span lengths from a Poisson distribution with $\lambda=3$ and replace these with a single mask token. We do not replace by a random token similar to BERT \cite{BERT} as early experiments showed it does not perform very well.

\subsection{Pretraining Objectives}
\label{Appendix:Pretraining-Obj}
The original BART pretraining involved a number of de-noising tasks including span masking, token deletion, sentence permutation, and document rotation \cite{bartlewis2020}. However, we argue in this work that these objectives are overly general and not specific for the dialogue domain. Here, we describe our proposed pretraining tasks:

\begin{itemize} 
   \item  \textbf{Masking pronouns}
    Conversations are usually rife with pronouns used for co-reference. In many cases. predicting the correct pronoun would require sufficient understanding of the dialogue context. We use a separate probability $P_{mask\_pronoun}=0.5$ of masking a specific pronoun.
    
    \item \textbf{Masking High-content tokens}
    While BART masking objective treats all tokens equally i.e all tokens are equally likely to be masked, we know that certain tokens are more relevant to a particular dialogue than other. Thus, here we choose to mask more salient tokens where salience is measured using TF-IDF weights. We start by computing TF-IDF weights over the whole dataset. Then for every input, we select the top 25\% weighted tokens and mask these with probability $p_{mask\_tfidf}=0.7$. This is somehow similar to PEGASUS \cite{zhang2020pegasus}, but here we mask tokens not sentences.

    \item \textbf{Masking Entities}
    \newcite{realm} showed that masking entities and dates could be helpful for IR tasks. We hypothesize that masking entities such as persons and locations can be particularly important for dialogues. Here we mask entities with probability $P_{mask\_entity}=0.7$. We use Spacy\footnote{https://spacy.io/} English NER model to detect entities.
    
\end{itemize}

% \begin{table*}[thb]
% \footnotesize
% \centering\renewcommand\cellalign{lc}
% \setcellgapes{3pt}\makegapedcells
% \begin{tabular}{|c|c|c|c|} \hline
% \textbf{Dialogue} & \textbf{Vanilla BART} & \textbf{Our Model} & \textbf{Reference} \\ \hline

% \makecell{
% \textbf{Sophia:} missed the tram... \\
% \textbf{Ella:} oh come on... \\
% \textbf{Sophia:} waiting for the next one I guess \\
% \textbf{Ella:} i guess...
% } & \makecell{Sophia missed\\ the tram.\\ Ella is waiting for\\ the next one.} & \makecell{Sophia missed\\ the tram and is waiting for\\ the next one.}
%     \\
    
% \hline\end{tabular}
%  \caption{Sample dialogues from SAMSum \cite{gilwa2019} that require reasoning for correct understanding/summarization.}
%     \label{App:tab-reasoning-examples}
% \end{table*}
\end{document}